\documentclass[sigplan,screen]{acmart}
\usepackage{graphicx}
\usepackage{subcaption}
\usepackage[export]{adjustbox} 

\AtBeginDocument{%
  }

\setcopyright{acmlicensed}
\copyrightyear{2018}
\acmYear{2018}
\acmDOI{XXXXXXX.XXXXXXX}

\acmConference[CODS'25]{Make sure to enter the correct
  conference title from your rights confirmation email}{Dec 17--20,
  2025}{Pune, India}
\acmISBN{978-1-4503-XXXX-X/2018/06}

\begin{document}

\title{AEGIS: An Agent for Extraction and Geographic Identification in Scholarly Proceedings}


\author{Om Vishesh}
\affiliation{%
  \institution{Central University of Jharkhand}
  \city{Ranchi}
  \country{India}}
\email{omvishesh123@gmail.com}

\author{Harshad Khadilkar}
\affiliation{%
  \institution{IIT Bombay}
  \city{Mumbai}
  \country{India}}
\email{harshadk@iitb.ac.in}


\author{Deepak Akkil}
\affiliation{%
  \institution{Emergence AI}
  \city{Helsinki}
  \country{Finland}}
\email{deepak@emergence.ai}

\renewcommand{\shortauthors}{Vishesh, Khadilkar, Akkil}

\begin{abstract}
  Keeping pace with the rapid growth of academia literature presents a significant challenge for researchers, funding bodies, and academic societies. To address the time-consuming manual effort required for scholarly discovery, we present a novel, fully automated system that transitions from data discovery to direct action. Our pipeline demonstrates how a specialized AI agent, `Agent-E', can be tasked with identifying papers from specific geographic regions within conference proceedings and then executing a Robotic Process Automation (RPA) to complete a predefined action, such as submitting a nomination form. We validated our system on 586 papers from five different conferences, where it successfully identified every target paper with a recall of 100\% and a near perfect accuracy of 99.4\% . This demonstration highlights the potential of task-oriented AI agents to not only filter information but also to actively participate in and accelerate the workflows of the academic community. 
  The demo is in the foot note\footnote{Demo Link : \url{https://youtu.be/cKkKZGaMo7o}}. 
\end{abstract}

\ccsdesc[500]{Applied computing~Digital libraries and archives}
\ccsdesc[500]{Information systems~Specialized information retrieval}

\keywords{Task-Oriented AI Agents, Scholarly Workflow Automation, Information Extraction}
\maketitle

\section{Introduction}

In this paper and associated demonstration, we describe a task-oriented AI agent for (1) extracting lists of accepted and/or published papers from a proceedings webpage, (2) analyzing the author lists for association with a specific geography (in this case, India), and (3) submitting the resulting papers for collation on the IKDD Premier Papers webpage\footnote{\url{https://ikdd.acm.org/premier-papers.php}}. Based on the open source Agent E \cite{abuelsaad2024agent} from Emergence AI, we show that a carefully designed pipeline is able to methodically process a large volume of information, analyze its relevance to the task, and perform pre-specified actions with very high precision and recall. 

We have several goals for sorting papers based on the authors' geographic association. Knowing where authors are based, where studies are conducted, and where collaborations occur enables richer bibliometric analyses, informs research policy, supports funding decisions, and sheds light on the global flow of knowledge. Further, we believe that indexing top-tier research output based on area of work enables stronger future research collaborations.

However, extracting this information is challenging since different proceedings and libraries have different formats for listing accepted and/or published papers. Some appear as HTML listings on conference websites or publisher pages, sometimes with direct links to PDFs, and other times linking to the article’s landing page on the publisher’s site. Even within a single paper, affiliations may be inconsistently formatted, spanning multiple lines, using superscripts and footnotes to link authors to institutions (e.g., A. Kumar$^1$), or listing multiple affiliations for a single author (e.g. C. Dickens$^{3,4}$). Sometimes affiliations omit explicit geographic information altogether, relying on readers’ prior knowledge. This diversity and ambiguity complicate reliable automated extraction.

This challenge of extracting geographic information from diverse, inconsistently formatted sources is not unique to academic proceedings. Similar issues arise across many domains where location data is embedded in semi-structured or unstructured text, such as patents, corporate filings, news reports, grant applications, and NGO publications. In each case, references to places and institutions vary widely in form, completeness, and clarity, often mixed with other metadata or narrative text, making methodically extracting this information very challenging.
\begin{figure*}[htbp]
    \centering 

    \begin{subfigure}[b]{0.32\textwidth}
        \centering
        \includegraphics[width=\textwidth, valign=t]{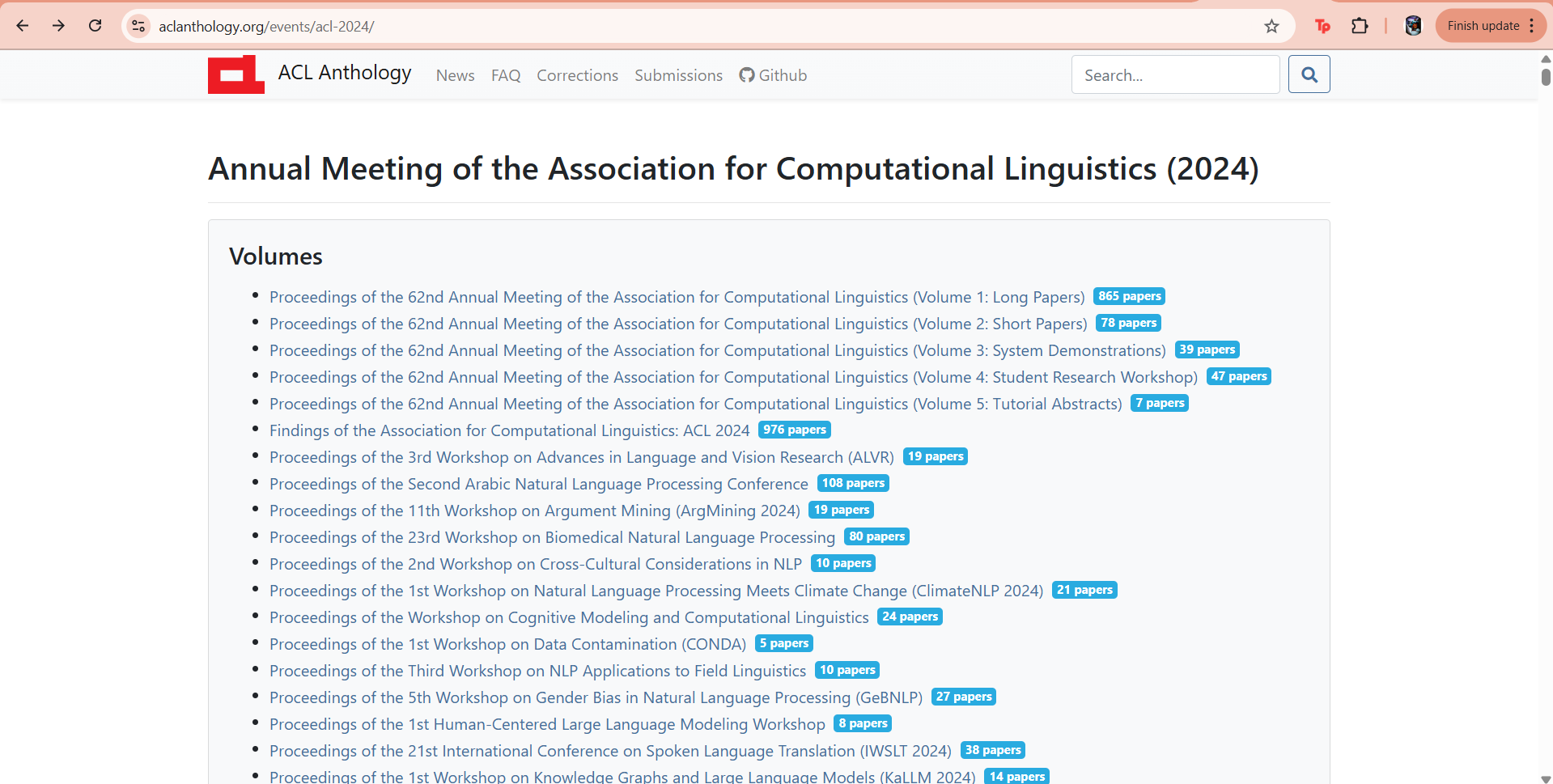}
        \caption{ACL 2024 proceedings page.}
        \label{fig:sub1}
    \end{subfigure}\hfill
    \begin{subfigure}[b]{0.32\textwidth}
        \centering
        \includegraphics[width=\textwidth, valign=t]{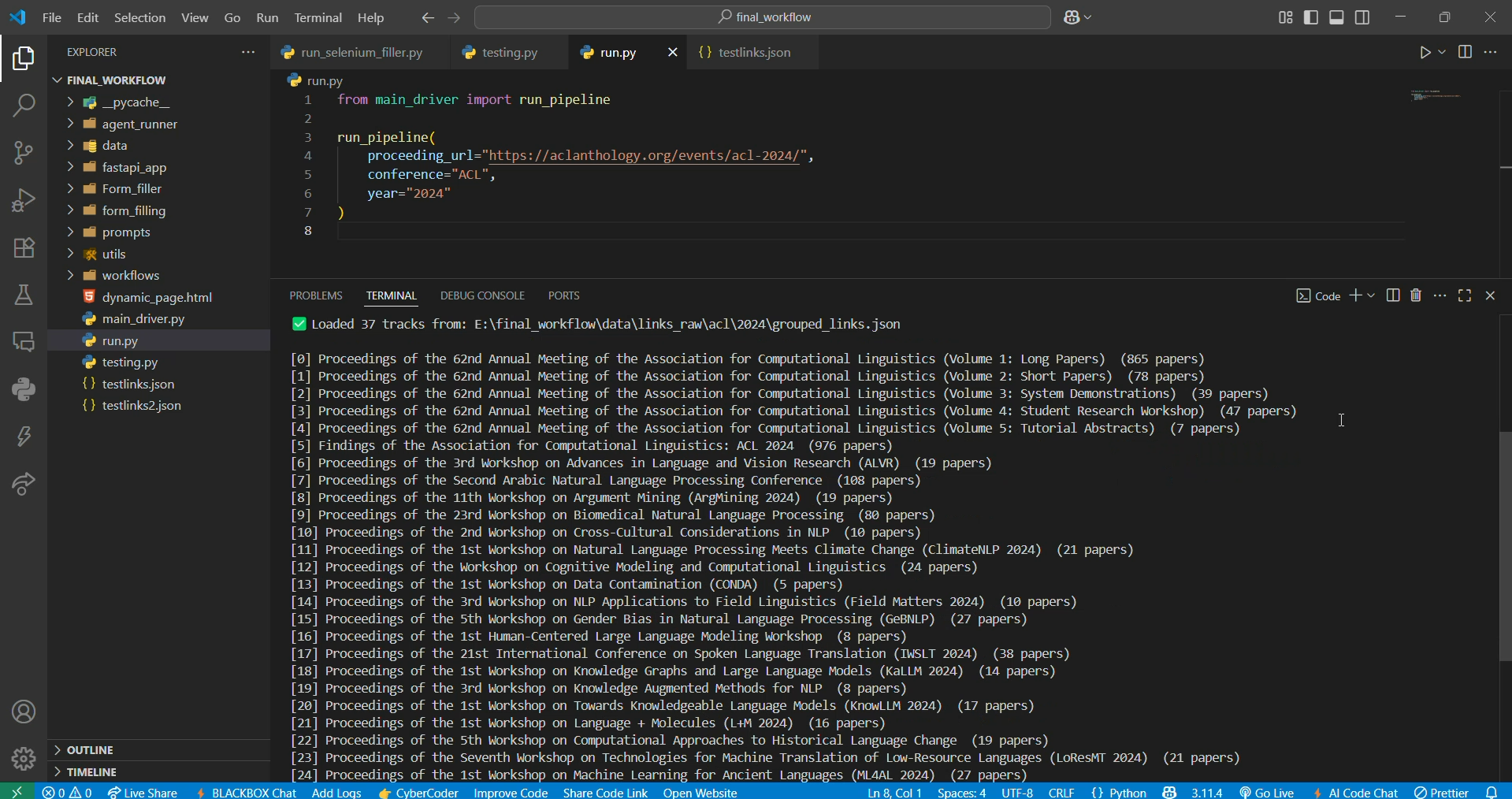}
        \caption{Loaded tracks for ACL.}
        \label{fig:sub2}
    \end{subfigure}\hfill
    \begin{subfigure}[b]{0.32\textwidth}
        \centering
        \includegraphics[width=\textwidth, valign=t]{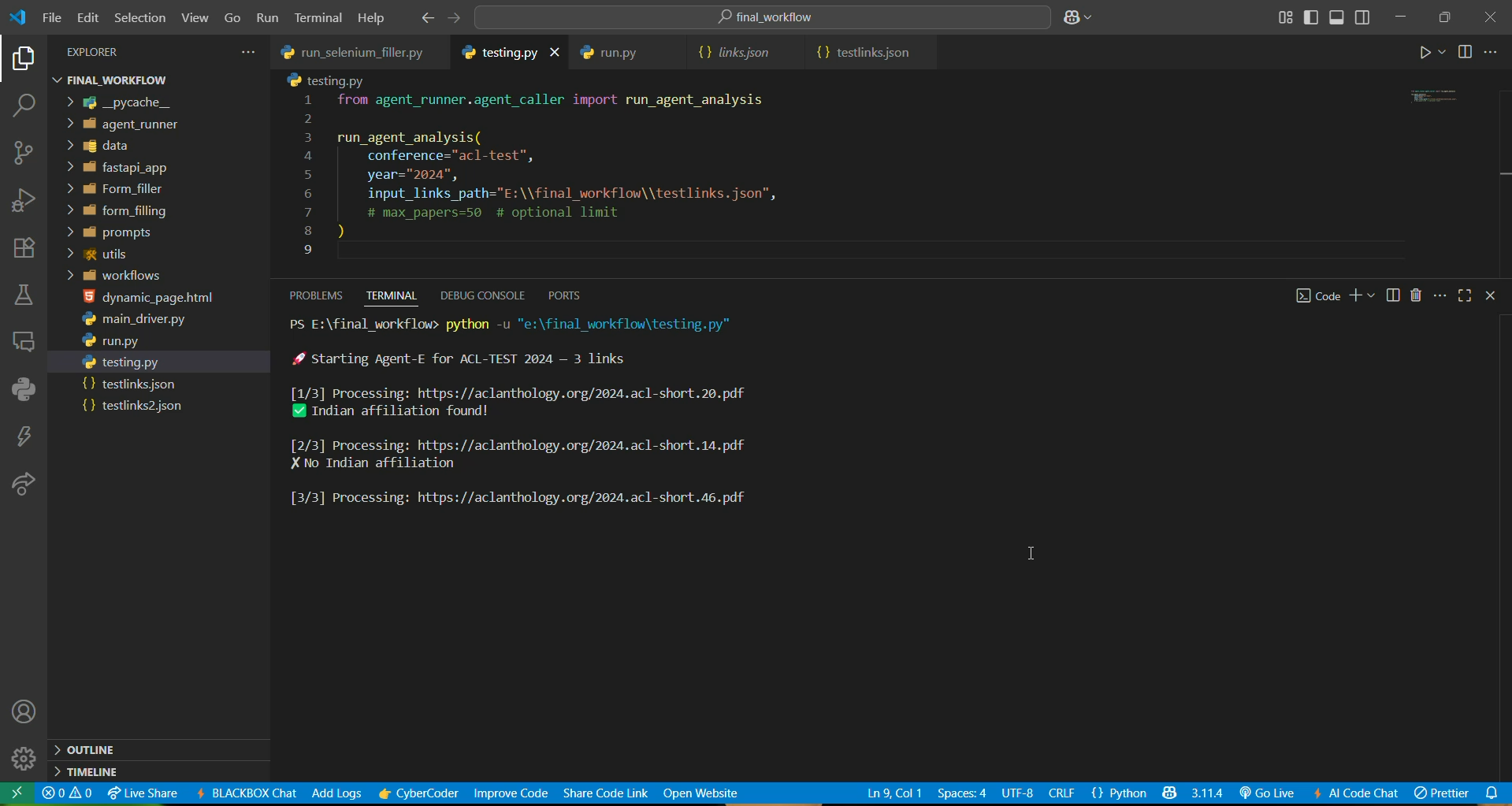}
        \caption{Agent processing the papers.}
        \label{fig:sub3}
    \end{subfigure}
    
    \vspace{0.5cm} 

    \begin{subfigure}[b]{0.32\textwidth}
        \centering
        \includegraphics[width=\textwidth, valign=t]{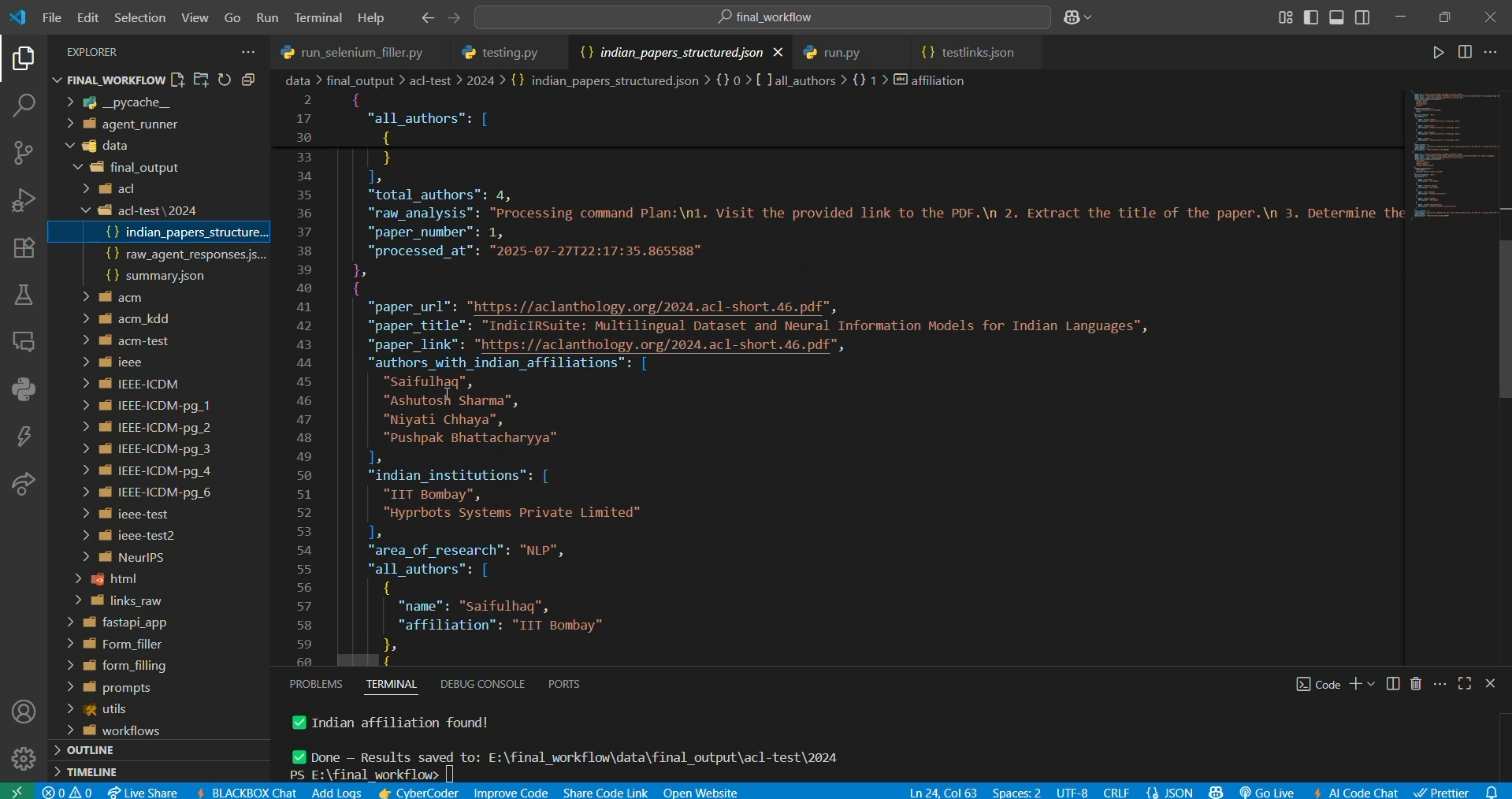}
        \caption{Extracted data of papers meeting the criteria.}
        \label{fig:sub4}
    \end{subfigure}\hfill
    \begin{subfigure}[b]{0.32\textwidth}
        \centering
        \includegraphics[width=\textwidth, valign=t]{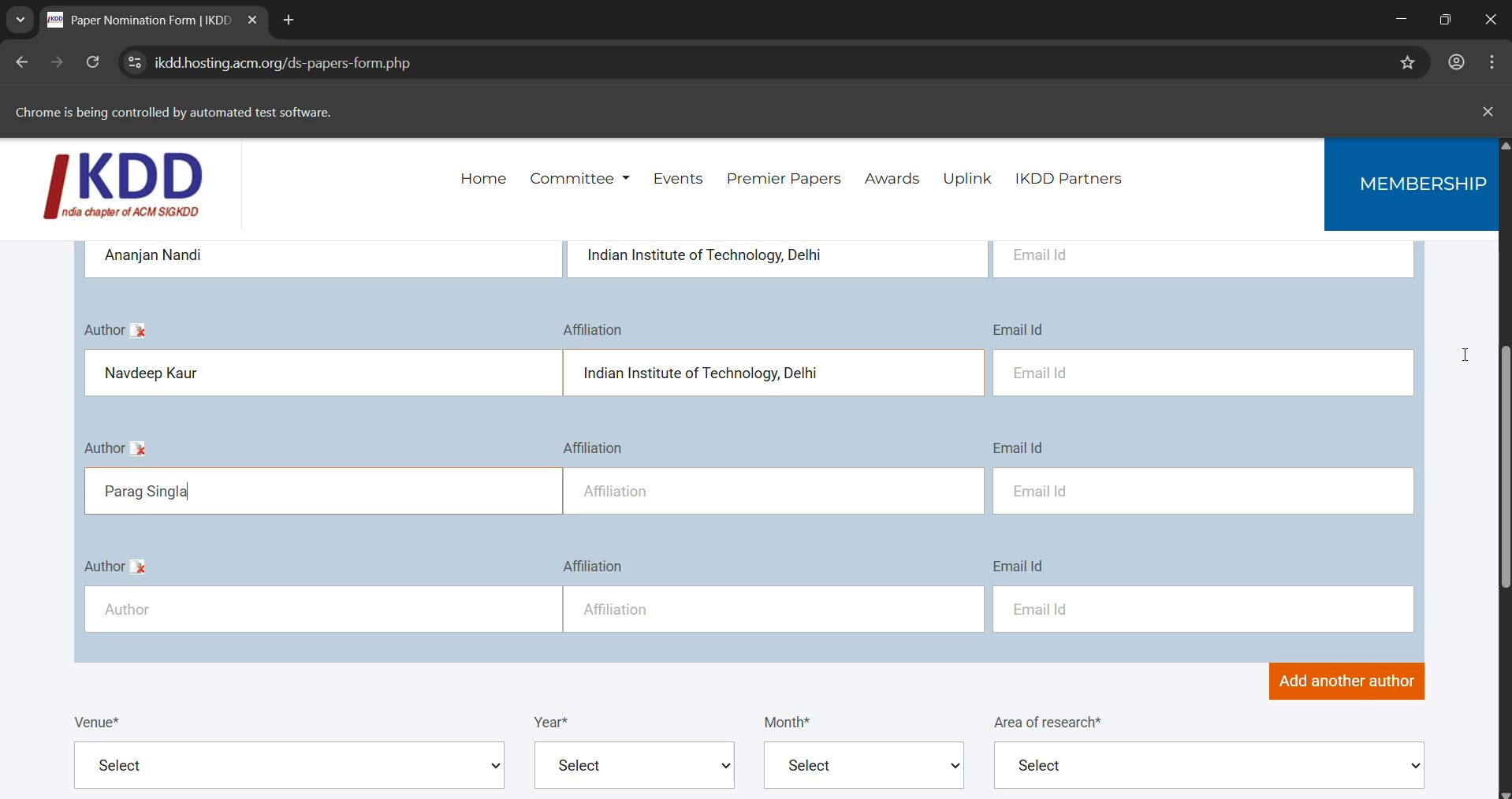}
        \caption{RPA filling the nomination form on IKDD website.}
        \label{fig:sub5}
    \end{subfigure}\hfill
    \begin{subfigure}[b]{0.32\textwidth}
        \centering
        \includegraphics[width=\textwidth, valign=t]{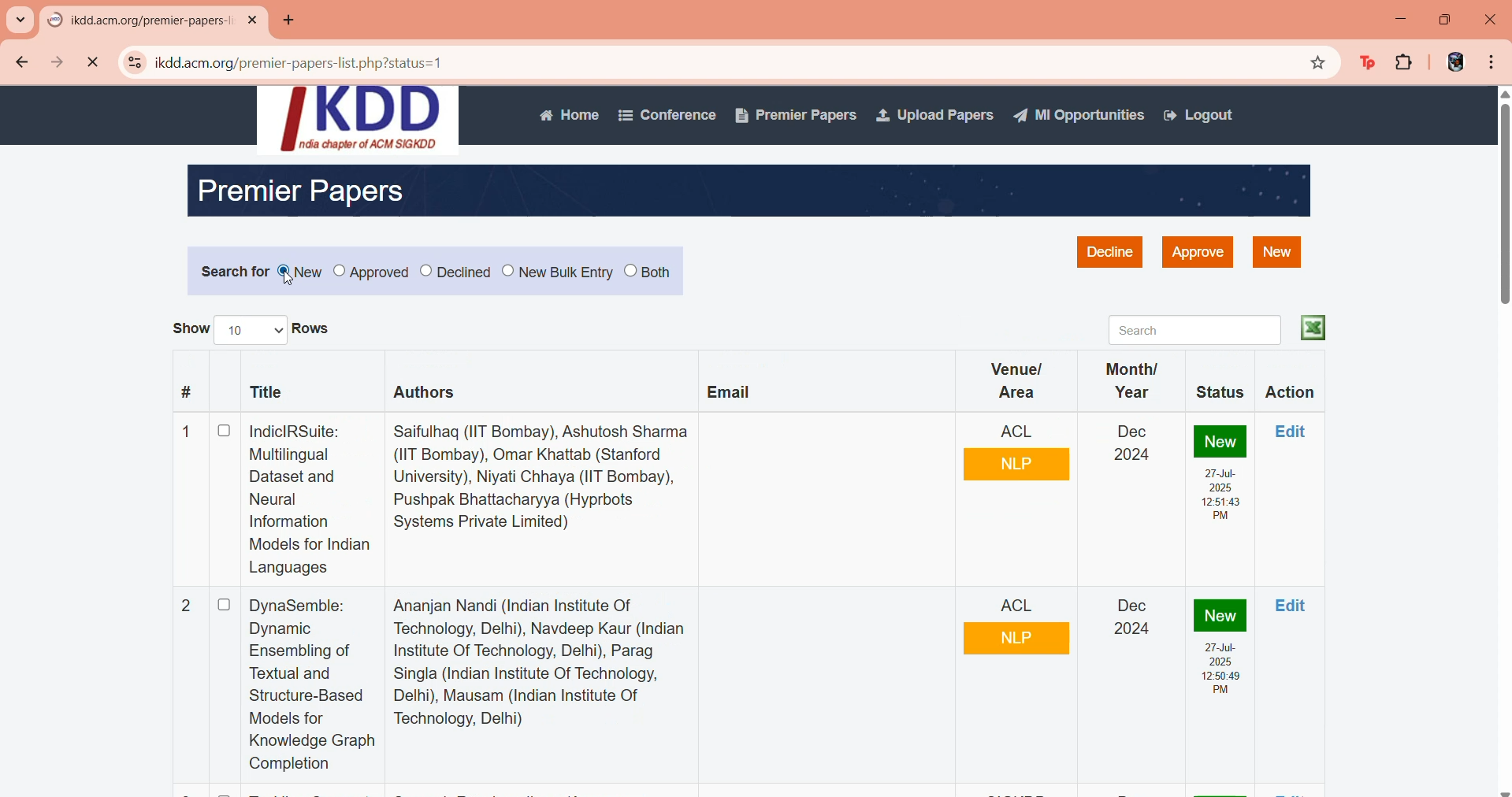}
        \caption{The submitted data showing up on the IKDD backend.}
        \label{fig:sub6}
    \end{subfigure}

    \caption{Screenshots of key stages in the workflow.}
    \label{fig:main_grid}
\end{figure*}

\section{Problem Description}

Consider an initial state where a conference has published its list of accepted papers, sorted by various tracks. Our goal is to start with a single piece of information (the url), and complete all the steps in Figure \ref{fig:main_grid} in a single workflow. This includes analyzing available tracks, checking their acceptability for ranking (e.g. workshops do not qualify), reading through the list of titles and authors (with or without affiliations), analyzing author affiliations for association with an Indian institute (for at least one author), generating entries for the IKDD Premier Papers page in the required format, and submitting the form for each eligible paper.

Unlike traditional rule-based systems, LLMs can adapt dynamically to diverse document formats, interpret ambiguous or incomplete affiliation data, and leverage background knowledge to resolve geographic references. Their ability to combine linguistic insight with reasoning presents a promising avenue toward scalable, accurate, and robust geographic metadata extraction in scholarly communications and beyond. While general-purpose browser-based AI agents like Operator \cite{cua2025} and Agent-E \cite{abuelsaad2024agent} can assist with certain sub-tasks, they are not designed to handle the full pipeline required for large-scale, long-form processing. Proceedings often contain hundreds or thousands of publications, which adds significant complexity and volume to the extraction task. Addressing this scale, along with the diversity and complexity of papers, calls for a custom, domain-specific solution that can effectively orchestrate the entire extraction process. Our approach integrates existing agents like Agent-E as components within a broader framework, enabling scalable and reliable information extraction from scholarly metadata.

\section{Proposed Methodology}
\begin{figure}
      \centering
      \includegraphics[width=0.95\linewidth]{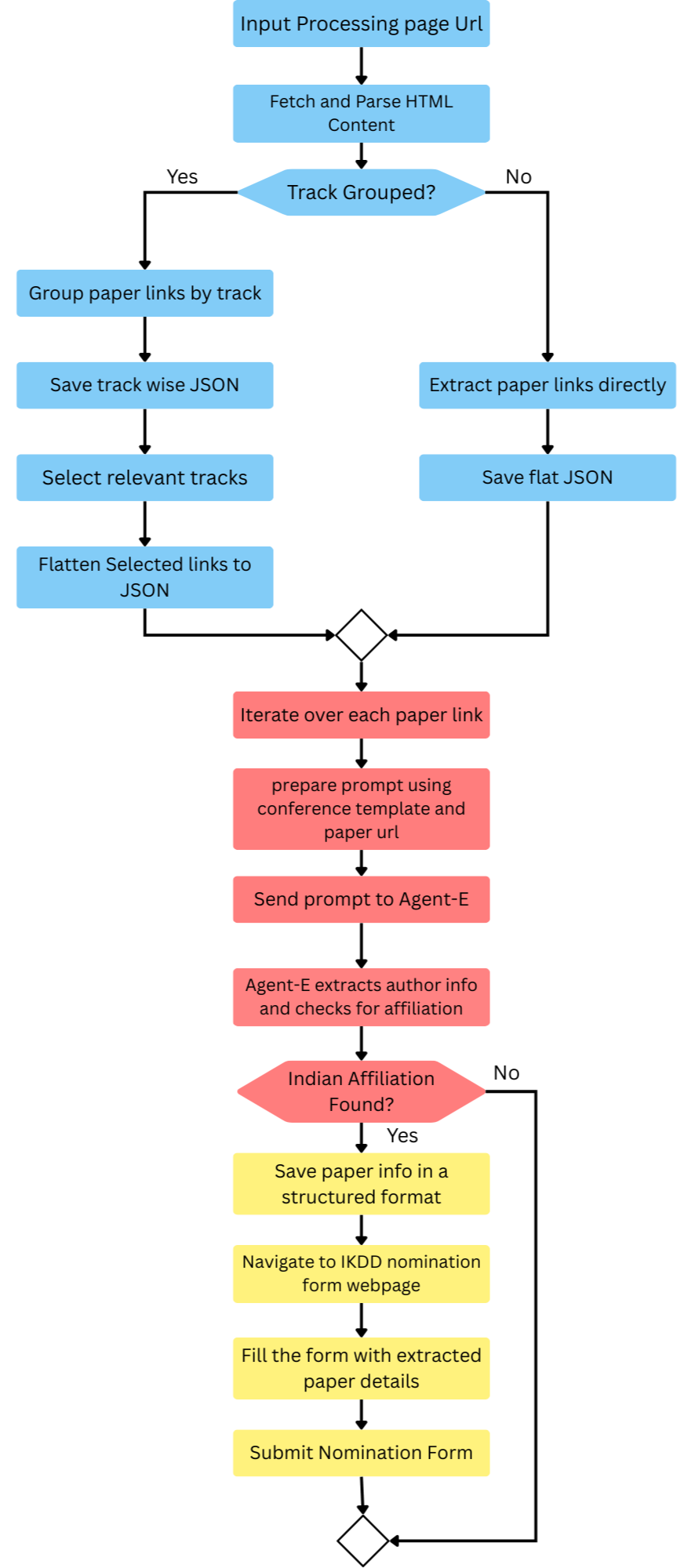}
      \caption{Full logical workflow of AEGIS.}
      \label{fig:flowchart}
      \Description{workflow of the agent}
  \end{figure}

Our proposed workflow and methodology is shown in Figure \ref{fig:flowchart}, which starts with data ingestion, moves to HTML parsing, hyperlink discovery, layout aware link normalization, dynamic prompt engineering and agent invocation, AI Response Parsing and Data Structuring, and finally automated Nomination via Robotic Process Automation. The last step involves filling the nomination form on the IKDD website based on the extracted information after identifying the papers that meet the criteria. Each step is described below.
  
\subsection{Data Ingestion and Source Acquisition}
To start the data collection pipeline, we created a primary execution script that defines the scope of a given task. This script takes some important parameters which includes the URL of a conference’s proceedings page, conference name, its year of publication, and a limit constraint optionally to restrict the number of papers to be processed or to continue a interrupted job.
To deal with web pages that load content dynamically, we added an automated web browser framework. This system automatically opens a browser, navigates to the destination URL, and waits intelligently for all of the dynamic page content to fully render. Once this render time has passed, we extract the final HTML source code and save it as a local file\footnote{One open item is to handle pagination of long paper lists, which is currently work in progress.}. We implemented this local caching strategy to ensure the reproducibility of our experiments and to minimize the network load on conference servers during development and testing.

\subsection{HTML Parsing and Hyperlink Discovery}
Now that we have the complete HTML source code required, we used a robust parsing library \cite{richardson2007beautiful} to convert the raw text into a structured parse tree. The tree represents the document’s hierarchical structure, making it programmatically easy to navigate.

To discover all the potential paper links, we systematically traversed this parse tree and extracted every hyperlink present on the page. This comprehensive search resulted in raw, unfiltered list of all URLs, which then served as the input for our normalization process.

\subsection{Layout-Aware Link Normalization}
For addressing the significant challenges of non-standardized formats across different conference websites, we designed and implemented a layout-aware normalization module. This module first analyses the conference identifier to automatically determine the structural layout of the proceedings page. Based on this, it dynamically selects one of the two specialized extraction strategies.

\textbf{For Flat-List Structure:} In cases where papers are presented in single, continuous list (common on platforms like the IEEE Xplore Digital library, NeurIPS), we applied a pattern-matching approach. We developed a configurable system that uses predefined filtering rules, including regular expressions, to precisely identify and extract valid paper URLs from other irrelevant links. This process also constructs absolute URLs from the relative paths and removes any duplicates.

\textbf{For Track-Based Structures:} For websites where papers are organized tracks or sessions (common for publishers like ACM and ACL), we developed a more sophisticated DOM traversal method. This method programmatically identifies the heading for each track and then intelligently navigates the parse tree’s parent and sibling relationships to reliably associate each heading with its corresponding block of paper links. The system then presents these tracks to the user via a simple command-line interface, allowing for the selection of relevant sessions.

Through this dual-strategy approach, our system robustly transforms structurally diverse source pages into a single, standardized, and clean list of paper URLs, ensuring consistent input for the subsequent phases of our research pipeline.

\begin{table*}[htbp]
  \centering
  \caption{Performance Metrics Across Different Conference Datasets}
  \label{tab:performance_metrics_standard}
  \resizebox{.75\textwidth}{!}{%
  \begin{tabular}{lcccccccc}
    \toprule
    \textbf{Conference} & \textbf{Total Papers} & \textbf{TP} & \textbf{TN} & \textbf{FP} & \textbf{FN} & \textbf{Accuracy} & \textbf{Precision} & \textbf{Recall} \\
    \midrule
    ACM SIGKDD  & 151 & 5  & 146 & 0 & 0 & 1.00 & 1.00 & 1.00 \\
    ACL         & 115 & 3  & 112 & 0 & 0 & 1.00 & 1.00 & 1.00 \\
    NeurIPS     & 100 & 5  & 95  & 0 & 0 & 1.00 & 1.00 & 1.00 \\
    IEEE ICDM   & 120 & 4  & 115 & 1 & 0 & 0.99 & 0.80 & 1.00 \\
    Custom Data & 100 & 20 & 78  & 2 & 0 & 0.98 & 0.90 & 1.00 \\
    \bottomrule
  \end{tabular}%
  }
\end{table*}

\subsection{Prompt Engineering and AI Agent Invocation}
For each normalized paper URL, we engineered a dynamic prompt generation system to instruct our AI agent `Agent-E' \cite{abuelsaad2024agent}. We created a library of prompt templates, each tailored to a specific conference series or publisher (one per library). The system first attempts to load a highly specific prompt template based on the publisher (e.g., a generic IEEE prompt). This ensures both precision and adaptability. The selected template is then populated with the paper’s unique URL to create the final instructional prompt.

This prompt is sent via a POST request to Agent-E’s REST API endpoint. Our system is designed to handle a streaming response, processing the agent’s output token by token as it is generated. This allows for real-time logging and immediate processing without waiting for the full analysis to complete, which is crucial when dealing with potentially long analysis times for full-text papers.

\subsection{AI Response Parsing and Data Structuring}
To convert the semi-structured text response from Agent-E into a reliable format, we developed a multi-stage parsing module. We first implemented a deterministic, rule-based parser that uses a series of regular expressions to extract simple key-value pairs like "Paper Title." For more complex data, such as the full author list, our system employs a robust JSON parsing mechanism. To handle potential formatting inconsistencies in the AI's output, this parser includes a custom fallback that uses a secondary regex-based approach to manually reconstruct the data if standard JSON decoding fails.

Following the initial parse, we added a crucial verification layer to ensure data quality and minimize false positives. Even if the agent reports a potential match, our system performs a final check to confirm that the extracted lists of authors and institutions are not empty or filled with null values (e.g., "None"). Only after passing this verification is the data considered a valid positive and structured into a final JSON object for the next stage of the pipeline.

\subsection{Nomination via Robotic Process Automation}
The final phase of our pipeline executes the automated submission of nomination forms using a Robotic Process Automation (RPA) approach. For each verified paper, the system launches a web automation session using the Selenium framework \cite{garcia2020survey}. It navigates to the target form and, to handle dynamic page elements, automatically adds the required number of author fields. To ensure reliable interaction, we implemented a JavaScript-based scrolling function that brings each element into view before it is clicked.

Once the form structure is correctly configured, the system populates all fields-including the paper title, author names, affiliation, and research area by mapping the structured data from our JSON file. Finally, it submits the form on its own and waits for a confirmation message to appear on the page, thereby verifying a successful transaction before proceeding to the next paper.

\subsection{Results}
We evaluated our agent's effectiveness by testing the pipeline on 586 papers from five diverse conference datasets, including a custom-curated collection designed to more rigorously test precision with a higher ratio of positive instances(20 out of 100). As detailed in Table 1, the system's performance was exceptionally high. It achieved perfect 100\% accuracy, precision and recall on three datasets (ACM SIGKDD, ACL and NeurIPS). On more challenging IEEE ICDM and custom datasets, it maintained 99\% accuracy, with only a few false positives leading to high precision scores of 0.80 and 0.90 respectively. These minor errors were attributed to ambiguous affiliation strings. For example in IEEE the agent misclassified the "Table of Contents" document available in the proceedings as Indian affiliated paper.

The most significant outcome of our evaluation is that the system achieved a perfect recall of 1.00 across all 586 papers, with zero false negatives recorded. For a discovery and nomination pipeline, the primary objective is to ensure no relevant paper is ever missed, making perfect recall the paramount metric. Our analysis confirms that the pipeline is a highly reliable tool for this task, successfully balancing high precision with the crucial goal of flawless recall.
\subsection{Conclusion}
In this paper, we presented a system that successfully automates the discovery and nomination of geographically-targeted research from academic proceedings. We have demonstrated that by integrating an intelligent AI agent with a robust data preparation and RPA pipeline, it is possible to bridge the gap between simple information extraction and meaningful, real-world task execution. Our evaluation confirmed the system's high efficacy, achieving near-perfect accuracy and, most critically, a perfect recall rate of 100\%, which validates our approach for its intended application where failing to identify a relevant paper is not an option. Future work will focus on expanding the system's capabilities by developing new parsing modules for a wider array of conference publishers and major journal repositories like SpringerLink and arXiv. We also aim to improve precision by refining the AI agent's ability to disambiguate complex affiliation strings, thereby reducing the few false positives observed. Ultimately, this work serves as a proof-of-concept for a new class of task oriented AI agents that can significantly reduce manual effort and accelerate scholarly discovery.

\begin{acks}
This work was carried out as part of the ACM IKDD Uplink Internship. We would also like to acknowledge ideas and discussion received from Dr Indrajit Bhattacharya and Prof Amit Nanavati.
\end{acks}

\bibliographystyle{ACM-Reference-Format}
\bibliography{refs.bib}

\end{document}